\pdfoutput=1
\documentclass[11pt]{article}

\usepackage[preprint]{acl}

\usepackage{times}
\usepackage{latexsym}
\usepackage{enumitem}

\usepackage[T1]{fontenc}
\usepackage[utf8]{inputenc}
\usepackage{xcolor}
\usepackage{hyperref}
\usepackage{booktabs}
\usepackage{multirow}
\usepackage{adjustbox}
\usepackage[capitalise]{cleveref}
\usepackage{enumitem}

\newcommand{\modelname}{{Arctic-Embed 2.0}}

\hypersetup{colorlinks=true,linkcolor=blue,urlcolor=blue}

\usepackage{microtype}

\usepackage{inconsolata}

\usepackage{graphicx}

\title{\modelname{}: Multilingual Retrieval Without Compromise}

\author{Puxuan Yu\thanks{Corresponding author: puxuan.yu@snowflake.com} \and Luke Merrick \and Gaurav Nuti \and Daniel Campos \\
        Snowflake Inc. \\}

\begin{document}
\maketitle
\setlist{nolistsep} % save space on lists, see: https://tex.stackexchange.com/questions/6081/reduce-space-between-enumerated-items

\begin{abstract}
This paper presents the training methodology of \modelname{}, a set of open-source text embedding models built for accurate and efficient multilingual retrieval. While prior works have suffered from degraded English retrieval quality, \modelname{} delivers competitive retrieval quality on multilingual and English-only benchmarks, and supports Matryoshka Representation Learning (MRL) for efficient embedding storage with significantly lower compressed quality degradation compared to alternatives. We detail the design and implementation, presenting several important open research questions that arose during model development. We conduct experiments exploring these research questions and include extensive discussion aimed at fostering further discussion in this field.
\end{abstract}

\section{Introduction}

Transformer-based embedding models have become a cornerstone of various information retrieval (IR) applications (e.g., search engines and retrieval-augmented generation systems). Although many efforts have focused on English-only retrieval~\citep{merrick_arctic-embed_2024,wang_improving_2024,günther2024jina,nussbaum2024nomic}, considerable efforts have also been directed toward developing multilingual embedding models~\citep{wang_multilingual_2024,zhang_mgte_2024,chen_bge_2024,sturua_jina-embeddings-v3_2024}. By learning to map queries and documents from multiple languages into a shared representation space, these multilingual text embedding models enable non-English monolingual retrieval as well as cross-lingual retrieval. 

\begin{figure}
    \centering
    \includegraphics[width=0.9\linewidth]{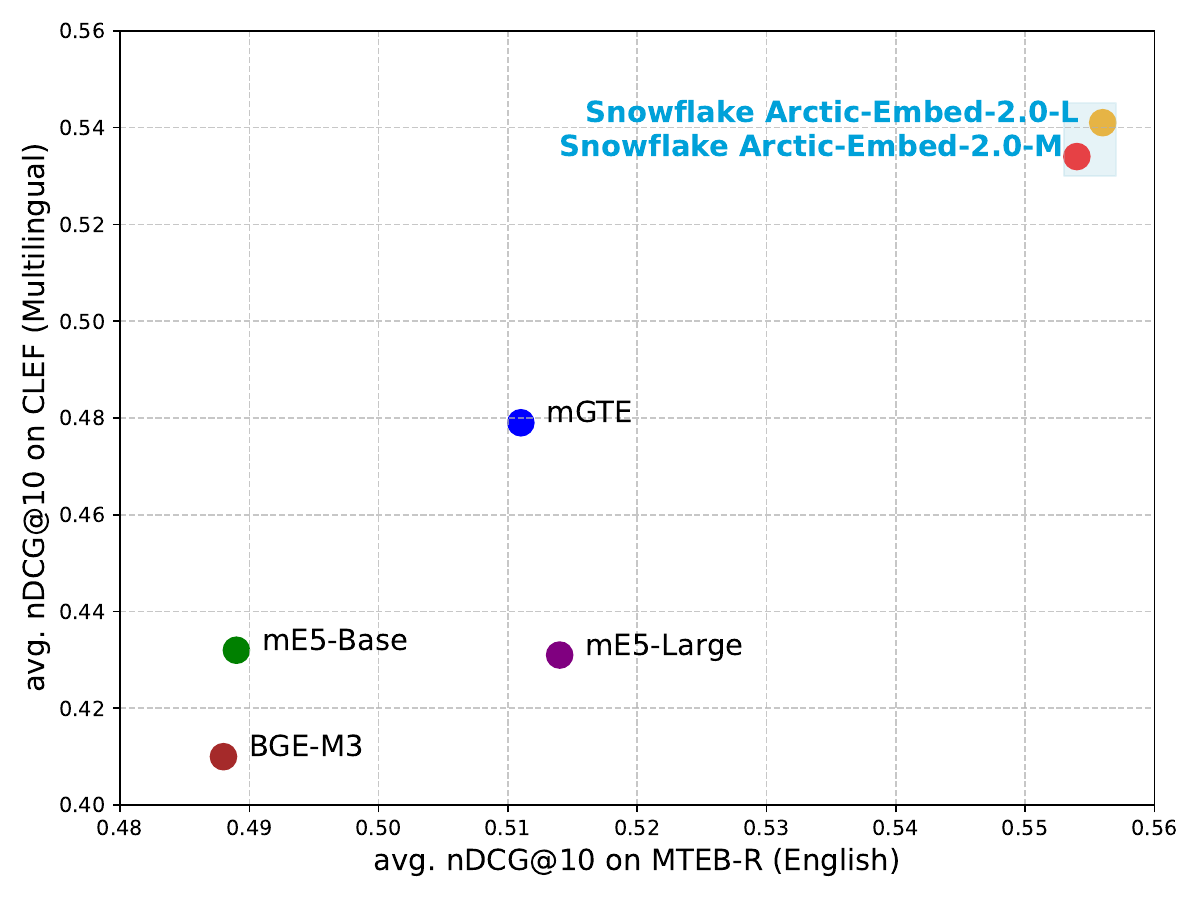}
    \caption{Single-vector dense retrieval performance of open-source multilingual embedding models with fewer than 1B parameters. Scores are average nDCG@10 on MTEB Retrieval~\citep{muennighoff2023mteb} and the subset of CLEF~\citep{clef} covering English, French, Spanish, Italian, and German.}
    \label{figure:teaser}
\end{figure} 

We develop \modelname{}\footnote{The open-source model weights are available under the Apache 2.0 License: 
\href{https://huggingface.co/Snowflake/snowflake-arctic-embed-m-v2.0}{snowflake-arctic-embed-m-v2.0} and \href{https://huggingface.co/Snowflake/snowflake-arctic-embed-l-v2.0}{snowflake-arctic-embed-l-v2.0}.} to deliver frontier retrieval quality while addressing two key limitations observed in current multilingual embedding models:

\textbf{Efficiency Losses}: Many state-of-the-art models that deliver high retrieval effectiveness require a large number of parameters and generate large embedding vectors \citep{wang_improving_2024,sturua_jina-embeddings-v3_2024}. These increase both computational and economic costs of dense retrieval, presenting challenges when dealing with large corpora.

\textbf{Compromised English Retrieval Quality}: It is common for multilingual models to underperform their English-only counterparts on English retrieval evaluations (e.g., MTEB Retrieval \citep{muennighoff2023mteb}). This trade-off has been a significant pain point in deploying multilingual systems.

% \modelname{} overcomes these limitations by offering a solution that improves retrieval performance across a wide range of applications, excelling not only at multilingual retrieval but also maintaining strong performance on English retrieval tasks. 
\modelname{} outperforms leading open-source alternatives, making it a highly versatile solution for both English and non-English contexts. Additionally, our two-stage approach to integrating Matryoshka Representation Learning (MRL)~\citep{kusupati_matryoshka_2022} drastically mitigates quality degradation during compression compared to other models supporting dimensionality reduction. 
% This enhanced compression efficiency enables \modelname{} to deliver superior performance at a fraction of the cost and complexity of competing models.
% In this report, we detail the methods and experiments behind \modelname{}, including training datasets, filtering techniques, and the implementation of pretraining and fine-tuning. Our objective was to achieve multilingual performance without compromising English quality, validating key assumptions along the way. Additionally, we present unexpected empirical findings, pose related research questions, and conduct further experiments to explore them.
Our contributions are as follows:
\begin{itemize}[leftmargin=*]
\item We introduce \modelname{}, models that achieve competitive retrieval quality on both English and multilingual benchmarks and support size-efficient embeddings via MRL.
\item We investigate potential causes of reduced English retrieval quality in prior multilingual models. We empirically refute the hypothesis that pretraining on distant languages harms English performance, and we propose alternative explanations for future investigation.
\item We reveal that pretrained checkpoint evaluations often fail to predict fine-tuned performance, highlighting the need for improved pretraining evaluation methods.
\item We show that while fine-tuning generally enhances cross-lingual transfer, contrastive pretraining in multilingual settings can lead to negative cross-lingual transfer.
\end{itemize}

% \section{Related Work}

% Cross-lingual and multilingual retrieval has been a central research area in natural language processing (NLP) and information retrieval (IR), driven by the need to support search and retrieval tasks across multiple languages. Multilingual retrieval deals with the notion of performing single-language retrieval, where the query and the document share a language, across many languages. Cross-lingual retrieval deals with retrieval where the query and documents do not have to share a language but can. 

% More recently, multilingual retrieval has leveraged multilingual extensions of pretrained language models, such as XLM-R~\citep{conneau_unsupervised_2020} and mBERT~\citep{Devlin2019BERTPO}. These models serve as the backbone for multilingual embedding models~\citep{wang_multilingual_2024,chen_bge_2024,zhang_mgte_2024,sturua_jina-embeddings-v3_2024}. While these models demonstrate impressive multilingual performance, they often encounter trade-offs, such as reduced performance in English retrieval tasks when optimized for multilingual scenarios~\citep{muennighoff2023mteb}.

% Training multilingual dense retrieval models requires careful consideration of both the model architecture and the training procedure. 
% In multilingual settings, the challenge is not only to learn a shared embedding space for multiple languages but also to ensure that the model can generalize effectively across different linguistic structures and domains~\citep{Artetxe_2019}.

\section{Methodology}
We follow a three-stage training framework inspired by prior works~\citep{merrick_arctic-embed_2024,wang_multilingual_2024,nussbaum2024nomic,chen_bge_2024,zhang_mgte_2024}: pretraining via masked language modeling, contrastive pretraining, and contrastive finetuning.

\subsection{Masked Language Modeling}
We use two open-source pretrained encoder models: \texttt{gte-multilingual-mlm-base} \citep{zhang_mgte_2024} for medium size and \texttt{bge-m3-retromae} \citep{chen_bge_2024} for large size.
% For our medium base model, we use \texttt{gte-multilingual-mlm-base}~\citep{zhang_mgte_2024}, which has 113M non-embedding parameters. For our large base model, we use the RetroMAE-adapted~\citep{xiao_retromae_2022} XLM-R base model, featuring 303M non-embedding parameters, developed as the first stage of the BGE-M3 model recipe~\citep{chen_bge_2024}. 
Both models employ the XLM-R tokenizer~\citep{conneau_unsupervised_2020}. Further details on the selection of these base models can be found in \Cref{appendix:select_base_models}.
\subsection{Contrastive Training Data}

% Another essential component in developing multilingual embedding models for retrieval is multilingual training data. 

\begin{figure*}
    \centering
    \includegraphics[width=0.8\textwidth]{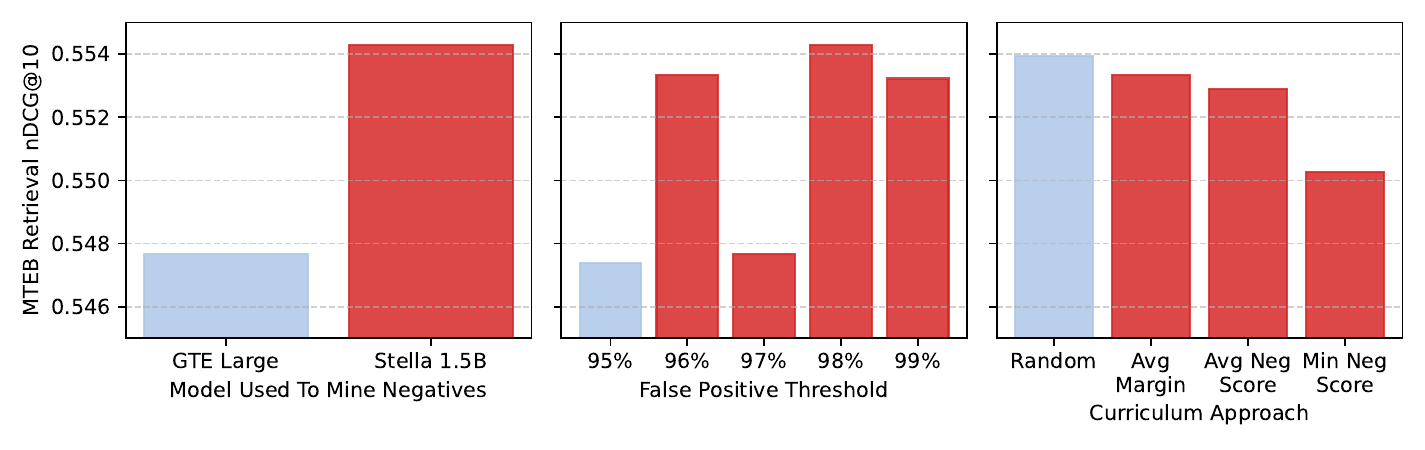}
    \vspace{-1em}
    \caption{Hard-negative mining ablation studies. A stronger teacher embedding model and well-tuned false-positive cutoff led to improved downstream performance, while a random order of examples performed just as well as various approaches to creating easy-to-hard curricula.}
    \vspace{-1em}
    \label{figure:negative-mining}
\end{figure*}

We report the details of our pretraining data in~\Cref{appendix:pretraining-data} due to space constraints. For finetuning data, we follow the data mix of~\citet{merrick_arctic-embed_2024} for English, adding MIRACL~\citep{zhang_miracl_2023} training set for high-quality multilingual training samples. We exclude Mr. Tydi~\citep{zhang_mr_2021} due to overlap with MIRACL but use all MIRACL languages (not just target ones), as we observe no negative transfer to retrieval in target languages.

\subsection{Methods of Data Filtering and Training}
We apply heuristic and consistency quality filters to English-only pretraining data following~\citet{merrick_arctic-embed_2024}. For multilingual pretraining data, we adopt retrieval-based consistency filtering as in several prior works~\citep{nussbaum2024nomic,jinav1,wang_improving_2024,dai2022promptagator}, using the small multilingual E5 model~\citep{wang_multilingual_2024} to embed queries and documents. Each dataset is partitioned into even shards of approximately 3 million query-document pairs, and pairs are filtered out if the pair's document ranks below rank 20 by vector similarity within all documents in its shard.

For contrastive training, we adopt the same approach as \citet{merrick_arctic-embed_2024} (see \Cref{appendix:hyperparams} for implementation details including training objectives, learning rates and schedules).

\subsection{Hard Negative Mining for Finetuning}\label{sec:finetune}

% ```python
% from matplotlib import rc
% rc('font',**{'family':'serif','serif':['Times']})

% from typing import NamedTuple

% import matplotlib.pyplot as plt

% class PlotInput(NamedTuple):
%     x: list[str]
%     y: list[float]
%     xlabel: str
%     rotate_xticks: bool = False

% plot_inputs = [
%     PlotInput(
%         xlabel = "Embeddings Used To Mine Negatives",
%         x = ['GTE Large', 'STELLA 1.5B'],
%         y = [0.54767, 0.5542805555555556]
%     ),
%     PlotInput(
%         xlabel = "Curriculum Approach",
%         x = ['Avg\nMargin', 'Avg Neg\nScore', 'Min Neg\nScore', 'Random'],
%         y = [0.5533394444444444, 0.5528805555555555, 0.5502761111111112, 0.55395],
%         rotate_xticks=False,
%     ),
%     PlotInput(
%         xlabel = "False Positive Threshold",
%         x = ['95%', '96%', '97%', '98%', '99%'],
%         y = [0.54739, 0.5533394444444444, 0.54767, 0.5542994444444445, 0.5532255555555555],
%     )
% ]

% ylim = (0.54, 0.56)
% ylabel = "MTEB Retrieval nDCG@10"
% xtick_label_rotation = 15
% fig, axes = plt.subplots(nrows=1, ncols=3, figsize=(9.5,3), sharey=True)
% for i, (plot_input, ax) in enumerate(zip(plot_inputs, axes)):
%     if i == 0:
%         ax.set_ylabel(ylabel)
%     ax.set_ylim(*ylim)
%     ax.set_xlabel(plot_input.xlabel)
%     ax.bar(plot_input.x, plot_input.y)
%     if plot_input.rotate_xticks:
%         for tick in ax.get_xticklabels():
%             tick.set_rotation(xtick_label_rotation)
%             tick.set_ha("right")

% fig.tight_layout()
% fig.savefig("./negative_mining_bar.pdf")
% ```

To select the most effective hard negatives, we adopt the strategy from NV Retriever~\citep{moreira_nv-retriever_2024}: documents scored as most relevant by a teacher embedding model are used as negatives, but any negative with a relevance score exceeding a specified percentage of the known-positive’s score is discarded as a potential false negative. We used \texttt{stella-en-1.5B-v5}\footnote{\scriptsize \url{https://huggingface.co/dunzhang/stella_en_1.5B_v5}} as the English teacher model, and \texttt{multilingual-e5-large} for multilingual data. Using \texttt{gte-large-en-v1.5} for comparison, we confirm~\citet{moreira_nv-retriever_2024}’s finding that stronger teacher models yield higher-quality fine-tuning datasets (\Cref{figure:negative-mining}, left). Rather than adhering to the $95\%$ false-positive filtering threshold suggested in prior work, however, we experimented with varying thresholds and observed improvement at higher thresholds (\Cref{figure:negative-mining}, middle).

% Additionally, we explored a curriculum learning approach to hard negative mining, similar to the method described as beneficial by~\citet{merrick_arctic-embed_2024}. In particular, we tried ordering the data with increasing negative hardness over the course of training using a variety of measurements of negative hardness (average margin between relevance score of negatives and relevance score of known-positive, average relevance score of negatives, and minimum relevance score of negatives). However, as shown in~\Cref{figure:negative-mining} (right), we found that a random ordering of the data actually led to comparable or better model quality than any curriculum approach.

We also explored curriculum learning for hard negative mining, inspired by~\citet{merrick_arctic-embed_2024}. Specifically, we ordered data by increasing negative hardness during training, using metrics like the average margin between relevance scores of negatives and known-positives, average negative relevance score, and minimum negative relevance score. However, as shown in \Cref{figure:negative-mining} (right), random data ordering produced comparable or better results than any curriculum-based approach.

\subsection{Matryoshka Representation Learning}

% While the modest model sizes of \modelname{} make low-latency, high-throughput inference straightforward and cost-effective on modern GPU hardware, achieving scalability and efficiency in downstream retrieval systems often requires optimizing the memory footprint of embedding vectors. This is because many computational costs in retrieval scale proportionally with the total number of bytes consumed by embeddings~\cite {aguerrebere2023similaritysearchblinkeye}.

% One effective approach for compressing embedding vectors into a smaller memory footprint with minimal degradation in retrieval quality is the combination of Matryoshka Representation Learning~\citep{kusupati_matryoshka_2022} with scalar quantization of the elements of the resulting vector~\citep{arctic1dot5blog}. \modelname{} uses this approach -- throughout both the pretraining and finetuning stages, we apply MRL loss at dimensionality 256 to both medium and large models, selecting this single embedding size to keep the distribution of individual embedding components more homogeneous and amenable to further compression via aggressive quantization. 

While the modest model sizes of our models enable inference with low latency and high throughput on modern GPU hardware, scalability in downstream retrieval systems often depends on optimizing the memory footprint of embedding vectors, since retrieval costs typically scale with the total memory consumed by embeddings~\cite{aguerrebere2023similaritysearchblinkeye}. \citet{arctic1dot5blog} showed that combining MRL~\citep{kusupati_matryoshka_2022} with scalar quantization is an effective method for compressing embeddings with minimal retrieval degradation. We emulate this approach, applying MRL loss during both pretraining and finetuning stages at a single truncated dimensionality of 256. This enables the medium and large models to achieve 3x and 4x compression, respectively, while ensuring a homogeneous distribution of components that facilitates aggressive quantization for further compression.

\section{Benchmarking}

\begin{table*}
\small
\centering
\begin{adjustbox}{width=0.75\textwidth}
\begin{tabular}{@{}l|ccc|cccc@{}}
\toprule
Model Name              & Multilingual?        & \#Params & Emb. Dim.            & MTEB-R                 & CLEF                 & MIRACL               & MIRACL-O             \\ \midrule
E5 Base v2               & no                   & 86M                           & 768                  & 0.502                & -                    & -                    & -                    \\
ME5 Base    & yes                  & 86M                           & 768                  & 0.489                & 0.432                & \underline{0.608}                & 0.509                \\
GTE Base En v1.5                & no                   & 113M                          & 768                  & 0.540                & -                    & -                    & -                    \\
GTE Multilingual Base & yes                  & 113M                          & 768                  & 0.511                & 0.479                & \textbf{0.621}                & 0.523                \\
Arctic-Embed 1.0-M            & no                   & 86M                           & 768                  & \underline{0.549}                & -                & -                & -                \\
\modelname{}-M   & yes                  & 113M                          & 768                  & \textbf{0.554}                & \textbf{0.534}                & 0.592                & \textbf{0.552}                \\
\quad{} + truncation   & yes                  & 113M                          & 256                  & \underline{0.549}                & \underline{0.522}                & 0.578                & \underline{0.545}                \\ \midrule
E5 Large v2               & no                   & 303M                          & 1024                 & 0.506                & -                    & -                    & -                    \\
ME5 Large   & yes                  & 303M                          & 1024                 & 0.514                & 0.431                & \underline{0.651}                & 0.540                \\
BGE Large En                     & no                   & 303M                          & 1024                 & 0.521                & -                    & -                    & -                    \\
BGE M3                  & yes                  & 303M                          & 1024                 & 0.488                & 0.410                & \textbf{0.678}                & \textbf{0.568}                \\
Arctic-Embed 1.0-L     & no                     &     303M                          & 1024                  & \textbf{0.560}                                                                         & - & - & - \\
\modelname{}-L    & yes                  & 303M                          & 1024                 & \underline{0.556}                & \textbf{0.541}                & 0.649                & \underline{0.558}                \\
\quad{}+ truncation    & yes                  & 303M                          & 256                  & 0.547                & \underline{0.530}                & 0.638                & 0.547                \\ 
\midrule
OpenAI Text Emb. 3 Large & yes & unknown & 3072 & 0.554* & 0.565 & 0.549* & - \\
\quad{} + truncation  & yes & unknown & 256 & 0.517* & 0.510 & - & - \\
Google Text Emb. 4 & no & 1.2B & 768 & 0.557* & - & - & - \\
\quad{} + truncation   & no & 1.2B & 256 & 0.524* & - & - & - \\
Google Text Emb. 4 Multilingual & yes & 1.2B & 768 & - & - & 0.562* & - \\
Voyage Multilingual 2 & yes & unknown & 1024 & - & 0.569 & - & - \\
\bottomrule
\end{tabular}
\end{adjustbox}
\vspace{-0.5em}
\caption{nDCG@10 performance of models in the evaluations, grouped by size. The best-performing model is highlighted in \textbf{bold}, while the second-best is \underline{underlined}. \#Params: the count of non-embedding parameters, except for Google's model, which only reports total parameters. Asterisks denote results from \citet{lee2024gecko}.} \label{table:main-results}
\vspace{-0.5em}
\end{table*}

\subsection{Evaluation Data Sets}

We evaluate English-only and multilingual retrieval using the widely adopted MTEB Retrieval benchmark~\citep{muennighoff2023mteb}, the MIRACL benchmark~\citep{zhang_miracl_2023}, and several languages from the CLEF 2000-2003 test suite~\citep{clef}. Since MIRACL is based exclusively on multilingual Wikipedia and its training dataset is widely used for embedding model training (including ours), it does not effectively assess a model’s ability to generalize beyond this domain. In contrast, the CLEF dataset, which lacks a training set and is derived from the news domain rather than Wikipedia, serves as a crucial tool for evaluating out-of-domain multilingual retrieval. Details of CLEF can be found in \Cref{appendix:clef}.

\subsection{Results}

The evaluation results, measured in nDCG@10, for MTEB-R, CLEF, MIRACL, and MIRACL-O (the subset of MIRACL languages that \textbf{o}verlap with our target languages -- English, French, Spanish, and German) are presented in Table~\ref{table:main-results}. 

%We run all open-source models ourselves while including some results published by~\citet{lee2024gecko} for closed-source models. We additionally run OpenAI's \texttt{text-embedding-3-large} model and Voyage AI's \texttt{voyage-multilingual-2} models on CLEF for additional points of reference.

Overall, our models achieve the best performance in their respective size categories on MTEB-R and CLEF, consistently outperforming same-sized competitors and delivering results comparable to flagship closed-source offerings. On MIRACL, our models are highly competitive, particularly for languages we specifically trained for. While models like mE5, mGTE, and BGE-M3 excel on MIRACL, their performance on CLEF is notably weaker compared to ours and closed-source offerings, suggesting the potential of \textbf{overfitting} to MIRACL or its Wikipedia-based domain.

Among models trained with MRL, \modelname{} pulls ahead as the clear leader when embeddings are truncated to 256 dimensions. In this setting, our medium size model outscores the best competitor (Google \texttt{text-embedding-004}) 0.549 to 0.524 on MTEB-R despite having far fewer model parameters, and our medium and large models retain 99\% and 98\% of the original MTEB-R scores, respectively -- substantially better than Google \texttt{text-embedding-004} (94\%)~\citep{lee2024gecko} and OpenAI Text Embedding 3 Large (93\%)~\citep{OpenAI_2024_blog}.
%Given the differing full-sized embedding dimensions, this truncation translates to a 3x improvement in retrieval efficiency for the medium model and a 4x improvement for the large model. 
We postulate that this stronger relative performance under truncation is a result of applying MRL to contrastive pretraining as well as contrastive finetuning, as \citet{lee2024gecko} indicate that MRL was only applied in the finetuning stage for Google's \texttt{text-embedding-004} model.

\section{Research Questions From The Journey}

Several empirical observations arose during the development of \modelname{} which lead to interesting and relevant research questions. Here we present two of these research questions which we explored through additional experimentation. Though our experiments shed some light on the situation, both questions remain open as important lines for future study.

\subsection{RQ1: Cross-lingual Transfer} \label{sec:rq1}

\textbf{How much does our large-scale contrastive pretraining benefit retrieval for languages not represented in the pretraining data?} Cross-lingual transfer (CLT) is a phenomenon wherein language-agnostic task knowledge is transferred from resource-rich source languages to target languages with limited or no resources. 
% CLT has been observed in tasks such as language modeling~\citep{conneau2019cross}, natural language inference~\citep{conneau2018xnli}, and task-oriented dialogue~\citep{schuster2019cross}. 
After observing strong scores across the full MIRACL benchmark (including on languages not covered by our contrastive pretraining), we focus on CLT in pretraining, though this phenomenon has also been studied in multilingual retrieval during the finetuning stage~\citep{zhang_toward_2023}.
% found that multilingual transformer models finetuned on relevance judgments in Thai were surprisingly effective in Arabic. 
% Inspired by studies investigating the mechanisms behind CLT in various NLP tasks~\citep{wu_beto_2019}, these authors attributed this improvement to code-switching in multilingual corpora and shared sub-words across languages.

\subsubsection{Experiments} We evaluate checkpoints of our medium model during contrastive pretraining at 2K-step intervals up to 10K steps, then at 10K-step intervals thereafter, assessing their performance on MIRACL. We evaluate in two ways: (1) direct evaluation of the checkpoint, and (2) finetune the checkpoint, then evaluate it. 
%The latter evaluation variation is important because in practice we care only about the final model's performance.

\begin{figure*}
    \centering
    \includegraphics[width=0.9\linewidth]{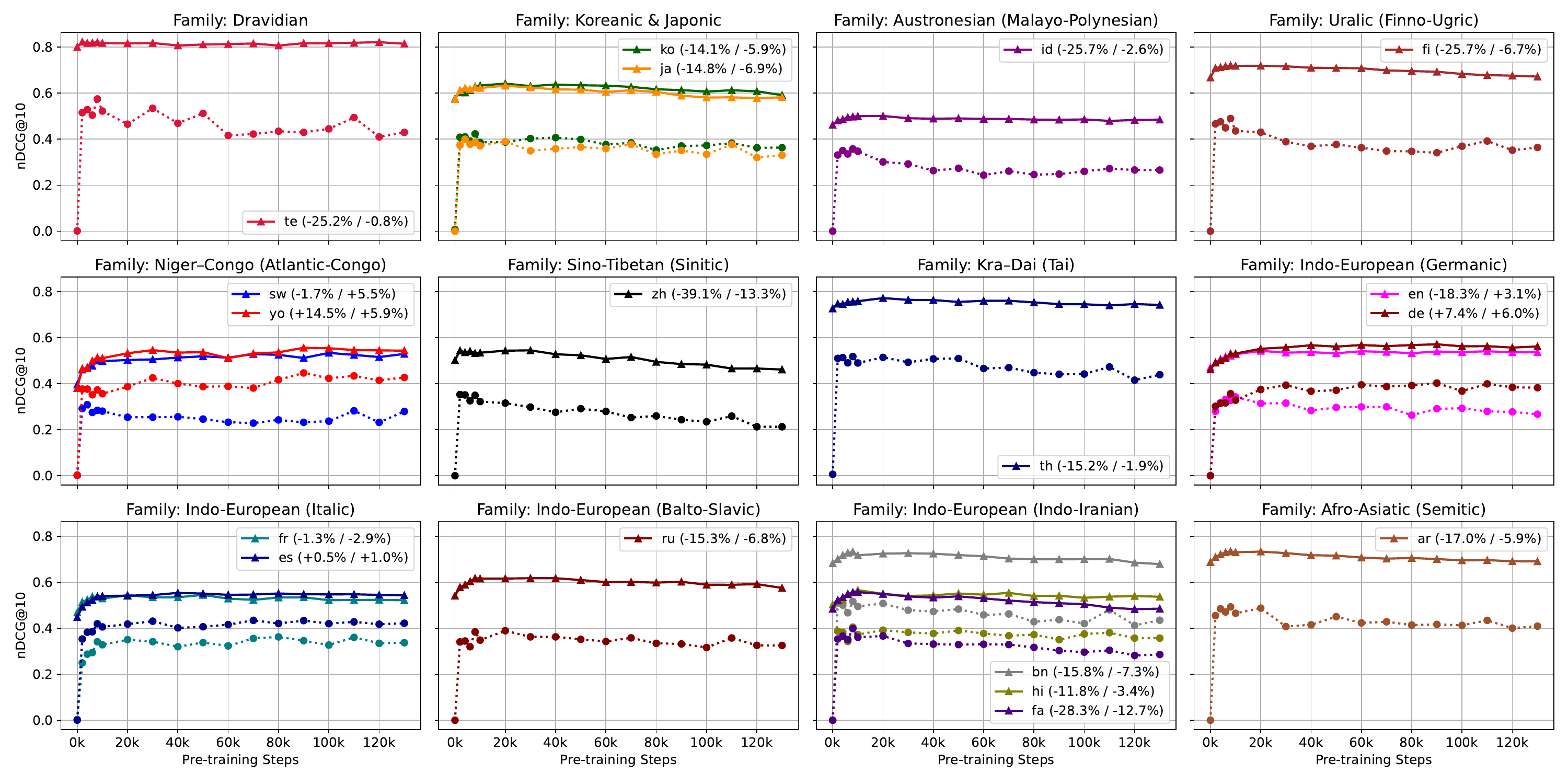}
   \vspace{-0.5em}
    \caption{MIRACL performance (in nDCG@10) at different points during contrastive pretraining. Languages are grouped by linguistic families provided by~\citet{zhang_miracl_2023}. Dotted lines represent non-finetuned runs, while solid lines represent finetuned runs. The relative improvement or deterioration of model performance at the end (130K steps) compared to the 8K-step checkpoint is reported for both non-finetuned and finetuned runs.}
    \vspace{-0.5em}
    \label{figure:miracl-along-pretraining}
\end{figure*} 

\subsubsection{Results} The evaluation results on MIRACL are shown in Figure~\ref{figure:miracl-along-pretraining}. Results with and without finetuning are represented by solid and dotted lines, respectively. 
% We observe that pretraining for 8K steps provides the most effective performance without finetuning for many languages, so we use this point as a benchmark to assess the impact of longer pretraining by calculating the relative improvement from this point to the end of pretraining. 
From this plot, we observe the following:

% \textbf{Evaluation before finetuning is misleading.} For example, evaluation without finetuning indicates that the 130K-step checkpoint performs 18.3\% worse than the 8K-step checkpoint on the English subset of MIRACL. However, after finetuning on the same data, the 130K-step checkpoint turns out 3.1\% better. This suggests that the complete train-finetune-evaluate cycle is essential for the accurate assessment of variations in pretraining (e.g., pretraining duration).

\textbf{Evaluation before finetuning can be misleading.} For instance, without finetuning, the 130K-step checkpoint appears 18.3\% worse than the 8K-step checkpoint on MIRACL’s English subset. After finetuning, however, it performs 3.1\% better.
%This highlights the necessity of the full train-finetune-evaluate cycle for accurately assessing pretraining variations (e.g., pretraining duration).

\textbf{We find little CLT in contrastive pretraining.} 
% Though previous studies show positive CLT effects in multilingual finetuning~\citep{zhang_toward_2023}, 
We observe negative trends in pre- and post-finetuned evaluation scores across most language families beyond those represented by the pretraining data. On evaluations with finetuning, the benefits of pretraining appear to peak within the first 10K steps, after which performance begins to decline for languages not represented by the pretraining data. 
% Surprisingly, despite the data mix containing French, we observed a 2.9\% post-finetuning performance drop on the French subset of MIRACL after the full pretraining run, suggesting other factors like data quality may play a larger role than either pretraining duration or CLT. 
We observe negative CLT effects particularly in Chinese (-13.3\%), Japanese (-6.9\%), Russian (-6.8\%), Finnish (-6.7\%), Korean (-5.9\%), and, surprisingly, French (-2.9\%), which makes up 13.8\% of the pretraining data (\Cref{figure:pretraining-data}). 
% The only exceptions are Niger-Congo languages like Swahili and Yoruba, which experience a positive CLT effect.

% \subsubsection{Alternative hypotheses} The reason for this negative result remains unclear. One possibility is that positive CLT observed during fine-tuning is primarily tested in-domain, where both the fine-tuning data in the source language and the evaluation data in the target language are sourced from Wikipedia in the same manner.
% It is likely that the positive CLT observed in prior studies is due primarily to in-domain patterns transferring across languages, while cross-lingual \textit{and} cross-domain knowledge transfer is more challenging. 
% Further investigation to validate these hypotheses is beyond the scope of this work, but we feel it represents a compelling direction for future research.
% We leave the validation of these hypotheses for future work.

% \px{It might be interesting to keep pursuing why -- i.e., why positive CLT on finetuning but not pretraining? Is it a data quality issue? But for now let's punt here and ``leave for future work :)''. }

% \px{Perform eval and finetune-then-eval on v1-pretraining -- running} 
% \px{Draw the same eval along pretraining graph for MTEB-R and CLEF. Why? On MIRACL we showed that pretraining for longer is bad, so we need some sort of justification.}

\subsection{RQ2: English Performance Gap}\label{sec:rq2}

\textbf{Why do many multilingual embedding models perform worse on English retrieval than English-only variants?}
% We observe across various embedding model families that multilingual embedding models often perform significantly worse on English retrieval tasks compared to their sister English-only models. 
As shown in Table~\ref{table:main-results}, transitioning from English-only to multilingual models results in drops of 1.3, 2.9, and 3.3 points on MTEB-R for E5 Base, GTE Base, and BGE Large, respectively, with several closed-source embedding models providers
% \footnote{
% At the time of writing, \href{https://cloud.google.com/vertex-ai/generative-ai/docs/learn/model-versions\#embeddings_stable_model_versions}{Google} offers both text-embedding-004 and text-multilingual-embedding-002, \href{https://docs.voyageai.com/docs/embeddings}{Voyage AI} offers voyage-3 and voyage-multilingual-2, and \href{https://docs.cohere.com/docs/models\#embed}{Cohere} offers embed-english-v3.0 and embed-multilingual-v3.0.
% }
also providing paired models with similar score gaps.\footnote{At the time of writing, Google, Voyage AI, and Cohere offer such English-multilingual model pairs.} However, despite this precedent, we observe strong English-language retrieval quality in our models. To understand why the degradation seen in other works appears absent in our results, we first conduct pilot experiments (\Cref{appendix:small-scale-ablation}) to confirm that we do not observe this language gap in our training.

\textbf{Initial hypothesis.} Unable to induce an English score gap in our training regimen (see \Cref{appendix:small-scale-ablation}), we look to the multilingual pretraining data used by other works to explain their English score gaps. Since our training data focus on European languages and we observe a negative transfer to certain non-European languages like Chinese in our RQ1 experiments, we hypothesize that certain languages may act as ``adversaries'' to English in retrieval tasks (i.e., training on these languages strongly diminishes English-language retrieval performance and vice versa).

\textbf{Experimental design.} To test this hypothesis, we paired English pretraining data with data from German, Spanish (controls), or Chinese (treatment). For each language, we sampled 600 batches (19.6M examples) from their respective corpora: web crawl for English, CC News for Spanish and German, and C-MTP~\citep{xiao_c-pack_2024} for Chinese. An English-only run was also trained for 16 epochs, while paired runs were trained for 8 epochs, totaling 314M samples per run. The English-only run was evaluated at the midpoint (``en'') to isolate \textit{data addition} effects, and at completion (``en+en'') to examine \textit{partial data replacement}. All runs were fine-tuned on the same data before evaluation.

\begin{figure}[h]
    \centering
    \includegraphics[width=0.9\linewidth]{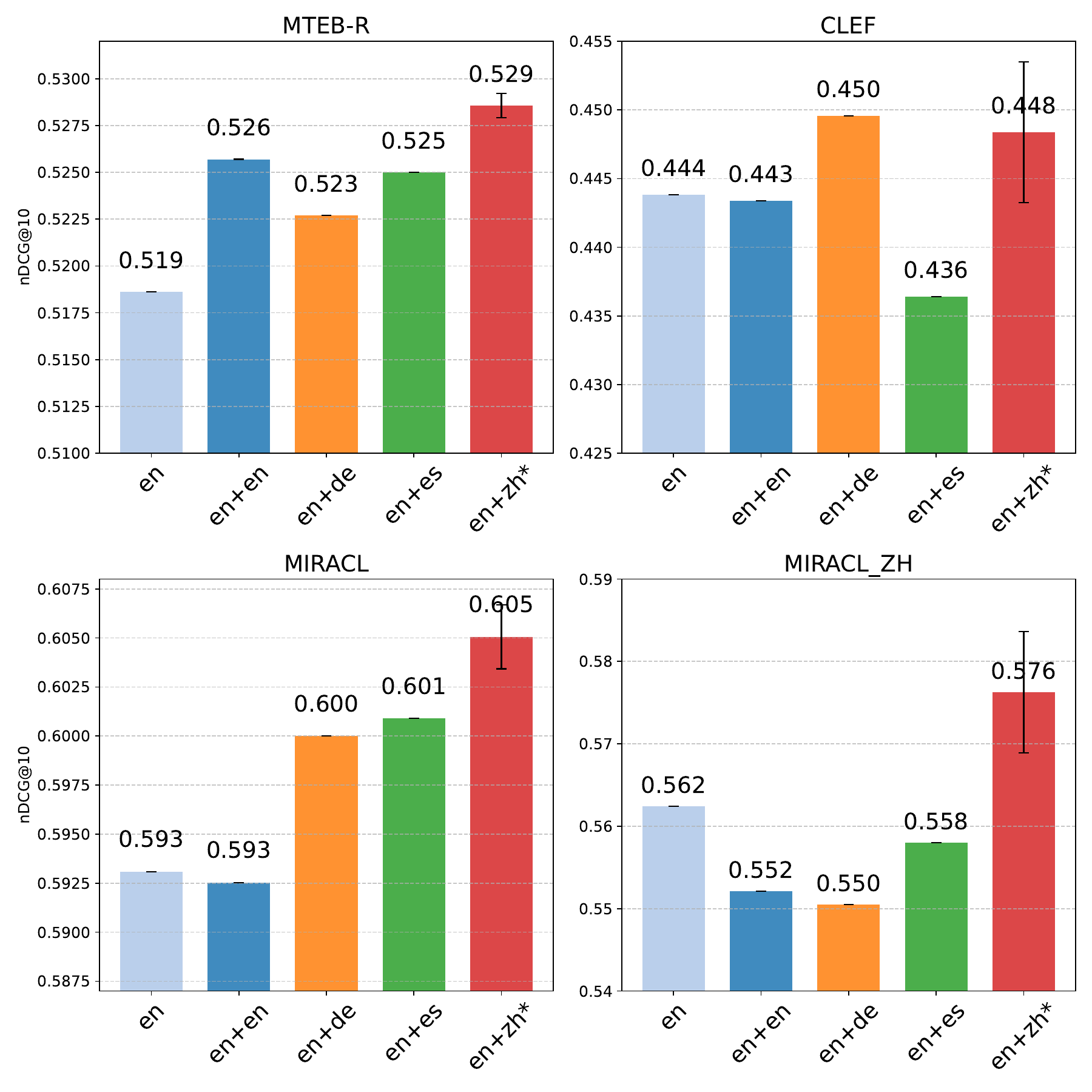}
    \vspace{-0.6em}
    \caption{The impact of adding equal amounts of English, German, Spanish, or Chinese data to the existing English pretraining baseline (``en'') on downstream retrieval performance. For Chinese data, error bars indicate the standard deviation across consistency filtering levels (top-\{1, 5, 10, 20, 30\} out of 3M), reflecting the effect of varying data quality.}
    \vspace{-0.5em}
    \label{figure:clt-with-chinese}
\end{figure}

\textbf{Results and analysis.} Figure~\ref{figure:clt-with-chinese} presents the outcomes of these experiments. Starting with the MTEB-R benchmark, it is evident that incorporating the Chinese pretraining data actually \textit{improves} English retrieval performance, whether as an addition to or a partial replacement of English data. This finding directly contradicts our initial hypothesis.
% that including pretraining data from languages distant from English negatively impacts English retrieval performance. 
Additionally, we see that this Chinese pretraining data outperforms Spanish and German CC News in improving retrieval across MTEB-R, MIRACL, and, in some cases, CLEF, which notably includes evaluation datasets for German and Spanish but not Chinese.
% This suggests that the filtered C-MTP dataset is of high quality, enabling positive cross-lingual transfer.
Finally, we note that these findings corroborate the trend observed in Figure~\ref{figure:miracl-along-pretraining}, where repeated training on English pretraining data primarily benefits English retrieval but provides limited improvement for other languages.

% \textbf{Alternative hypotheses.} One alternative hypothesis inspired by our results is that data quality plays a more significant role than language, with lower-quality multilingual training data explaining much of the gap experienced by prior works. A second alternative concerns model capacity -- perhaps it isn't the adversity of specific languages but the total number of languages trained simultaneously (exceeding a model’s capacity for learning) which could be the cause of the gap (with performance on English marginalized in favor of slightly improved performance on many other languages at once).

\textbf{Alternative hypotheses.} One hypothesis suggested by our results is that data quality plays a more critical role than language itself, with lower-quality multilingual training data accounting for much of the performance gap observed in prior works. Another hypothesis relates to model capacity: the issue may not stem from specific languages but rather from the total number of languages trained simultaneously, potentially exceeding the model’s capacity. This could lead to a trade-off where English performance is marginalized to achieve slight improvements across many other languages.

\subsection{Reflections On The Journey}

To summarize some key findings from our model development process, probing experiments, and our takes on interesting and important future directions:

\textbf{Data quality matters more than quantity.} We follow the advice of \citet{merrick_arctic-embed_2024} to emphasize data quality, deliberately rejecting lower-quality multilingual training data sources, performing retrieval-based consistency filtering, and carefully mining the best negative examples possible for fine-tuning. Though we do not extensively study lower-quality data, we rule out several other possible causes of lower retrieval performance observed in other works, and so we hypothesize that it is this focus on quality that explains why we do not observe degradation in English retrieval performance. In other words, \textbf{the English score gap observed in other multilingual models may simply reflect the challenge of acquiring high-quality retrieval training data in certain non-English languages}. Numerous empirical results from this paper lend credence to this quality-centric view, including our finetuning ablations in \Cref{sec:finetune} and the strong results across languages under a reduced training budget in \Cref{appendix:small-scale-ablation}.

\textbf{No clear formula for successful cross-lingual transfer in multilingual retrieval models.} In this work, we take a step toward improving the generalization of multilingual embedding models to unseen languages and domains, particularly by evaluating non-English retrieval beyond Wikipedia using CLEF~\citep{clef}. While we demonstrate the potential for multilingual training to enhance existing benchmark scores (e.g., better MTEB Retrieval scores with multilingual approaches in \Cref{appendix:small-scale-ablation}) and expand language support without penalty (e.g., overall benchmark improvements from adding Chinese in \Cref{figure:clt-with-chinese}), the actual process of model development is constrained by several factors: data quality (which is challenging to quantify), the availability of out-of-domain retrieval evaluation benchmarks in more languages, and the need to avoid exceeding model capacity (a concept still not fully understood). Faced with these limitations and uncertainties, incrementally and carefully incorporating non-English data into a proven English-language training mix has turned out to be the most effective strategy available for training useful multilingual embedding models.

\textbf{Model ``knowledge'' and task-calibration are both important yet possibly orthogonal.} As evidenced by the declining un-finetuned English retrieval scores in~\Cref{figure:miracl-along-pretraining}, it is entirely possible for pretraining to show the embedding model hundreds of millions more highly-filtered query-document pairs yet actually induce a \textit{degradation} in downstream retrieval after a certain point (20K steps out of 130K). Add on the finetuning step, however, and we find the performance trend reversed, with an extended pretraining regimen responsible for driving a 3\% \textit{increase} to the final nDCG@10 score! This flip-flop not only highlights the importance of measuring final downstream performance, but also demonstrates the intriguing possibility of some amount of useful knowledge being imparted ``silently'' into the model by contrastive training (somewhat analogously to how non-contrastive MLM pretraining improves the language model without making it fit for retrieval out-of-the-box). In hindsight, it appears that the success of \modelname{} may stem from both giving the model a substantial mount of ``knowledge'' through large-scale MLM and contrastive pretraining steps and from carefully ``recalibrating'' the model with the best and most denoised positive and hard-negative examples.

% Just as Statistical Learning Theory has studied the connections between surrogate loss functions and true cost functions through the concepts of calibration and consistency \citep{Steinwart2007HowTC}, perhaps the time has come for our field to study the ``calibration'' and ``consistency'' between various training regimens and the ultimate goal of quality retrieval.

% \textbf{Understanding the gap between two stages of training.} We found pre-finetuning metrics (e.g., evaluating pretrained checkpoints on retrieval benchmarks as in~\Cref{figure:miracl-along-pretraining}) not often indicative of the pretraining qualities. 

\section{Conclusion}

In addition to detailing the training process for \modelname{}, we investigate linguistic transfer in embedding models, revealing that prolonged contrastive pretraining does not always enhance cross-lingual transfer, though high-quality pretraining data from languages distant to English can surprisingly do so in some settings.
We further discuss how these experiments also uncover concrete evidence of the finetuning step of training ``reversing'' the negative impact of prolonged pretraining on downstream retrieval performance, a surprising result that indicates new direction for future scientific inquiry.

\clearpage{}

\bibliography{references, sample-base, custom}

\clearpage{}

\appendix

\section{Selection of Base Models}\label{appendix:select_base_models}

We aim to develop two model sizes to balance efficiency and effectiveness trade-offs. Based on pilot experiments with small-scale data, we selected the MLM-pretrained m-GTE checkpoint~\citep{zhang_mgte_2024} to initialize our medium model and the RetroMAE-pretrained BGE-M3 checkpoint~\citep{chen_bge_2024} for the large model. The m-GTE architecture leverages unpadding and xFormers acceleration~\citep{xFormers2022} for greater efficiency, while the BGE-M3 model benefits from RetroMAE pretraining~\citep{xiao_retromae_2022}, a retrieval-focused objective critical to its performance.

Both checkpoints use the XLM-R tokenizer. We also tested the newer Llama3 tokenizer~\citep{dubey2024llama}, designed for multilingual large language models. To accommodate this, we randomly reinitialized model weights, including the embedding layer, and pretrained both models directly without MLM pretraining. However, the Llama3 tokenizer showed no significant effectiveness gains and added efficiency overhead.

\section{Pretraining Data Breakdown} \label{appendix:pretraining-data}

Guided by end-user applications, we focused on European languages: English, French, Spanish, German, Italian, Portuguese, and Polish. For English data, we followed Arctic-Embed~\citep{merrick_arctic-embed_2024}. For multilingual data, we used mC4~\citep{habernal_c4corpus_2016}, CC News (treating page titles as queries and bodies as documents), and multilingual Wikipedia (titles and section headings concatenated as queries, section texts as documents) following~\citet{wang_multilingual_2024}. NLLB~\citep{team_no_2022} was excluded due to its limited resemblance to query-document tasks and negligible empirical benefit in early tests. For mC4, CC News, and mWiki, we subsetted to our target languages (including English).

\begin{figure}[h]
    \centering
    \includegraphics[width=\linewidth]{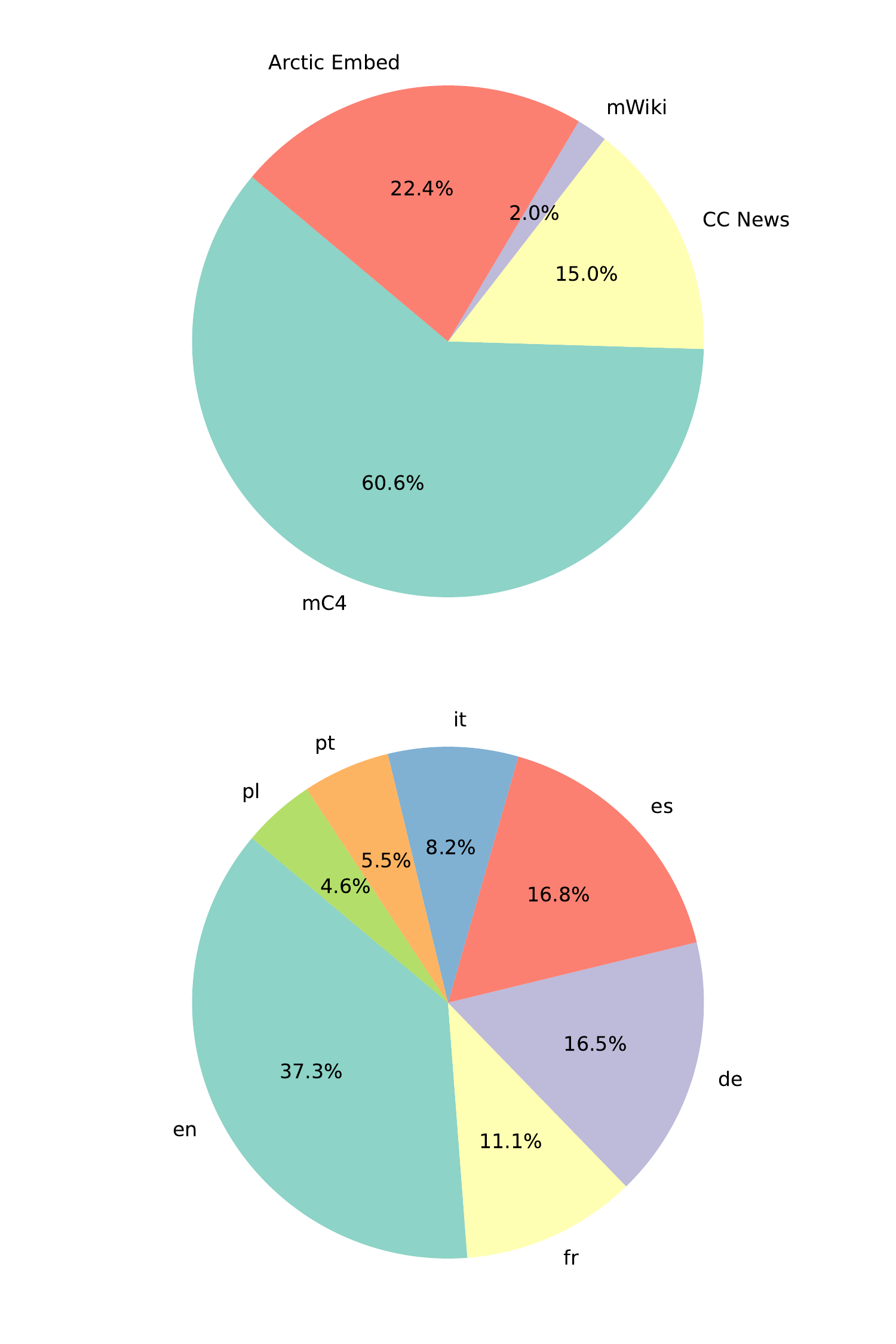}
    \caption{Breakdown of 1.41B contrastive pretraining samples by data source (top) and language (bottom).}
    \label{figure:pretraining-data}
\end{figure}

We apply top-20-in-3M-shard retrieval-based consistency filtering to each language subset of each dataset, resulting in approximately 1.41 billion unsupervised query-document pairs. A detailed breakdown of this combined dataset by source and language is shown in~\Cref{figure:pretraining-data}.

\section{Implementation Details} \label{appendix:hyperparams}

For pretraining, we use the standard InfoNCE contrastive loss~\citep{Oord2018RepresentationLW} with a temperature $\tau = 0.02$ as our contrastive pretraining objective. We use random in-batch negatives and follow the approach of~\citet{nussbaum2024nomic} and~\citet{merrick_arctic-embed_2024}, sampling all training mini-batches from a single data source at a time. For multilingual sources, different-language subsets are treated as distinct sources during batch construction. We use a batch size of 32,768, a maximum query length of 32, and a maximum document length of 256. We use peak learning rates of 3e-5 and 1e-4 for the large and medium models, respectively, following a linear warmup-stable-decay (WSD) schedule~\citep{hu2024wsd} for 3 epochs. To accommodate the large batch size and dataset scale, we employ activation checkpointing and use 32 H100 GPUs in a distributed data-parallel training setup.

In the finetuning stage, we train using the same InfoNCE loss function but rely on smaller, high-quality datasets and carefully curated negatives instead of random in-batch negatives. We also extend the maximum sequence length for queries and documents to 512 tokens, adjusting the batch size to 256 sets of 1 query, 1 positive doc, and 10 negative docs, changing the learning rate to 1e-5 and 5e-6 for medium and large models, respectively, and adjusting our WSD learning rate schedule to have no warmup and perform linear decay for 6,000 out of a total of 9,342 steps. 

\section{CLEF Dataset Details} \label{appendix:clef}

\begin{table}
\centering
\begin{adjustbox}{width=0.72\columnwidth}
\begin{tabular}{l|cccc}
\toprule
Language     & \multicolumn{1}{c}{\# Q} & \multicolumn{1}{c}{\# D} & \multicolumn{1}{c}{\# Rels} & \multicolumn{1}{c}{\# Rel/Q} \\ \midrule
English & 246                      & 113,005                  & 4,769                       & 19.4                         \\
French  & 185                      & 129,689                  & 3,022                       & 16.3                         \\
Italian & 176                      & 144,040                  & 2,626                       & 14.9                         \\
German  & 184                      & 153,496                  & 3,066                       & 16.7                         \\
Spanish & 156                      & 452,027                  & 5,759                       & 36.9                        \\ \bottomrule
\end{tabular}
\end{adjustbox}
\caption{Statistics of the CLEF evaluation datasets: number of queries (\# Q), corpus size (\# D), number of relevance judgments (\# Rels), and average annotations per query (\# Rel/Q).} \label{table:clef}
\end{table}

The statistics of the CLEF datasets used to evaluate the out-of-domain generalizability of multilingual models are reported in~\Cref{table:clef}. These datasets have been widely adopted in the literature on non-English monolingual retrieval~\citep{huang_improving_2023,huang_language_2024} and cross-lingual retrieval~\citep{yu_study_2020,yu_cross-lingual_2021,nair_blade_2023} as a reliable benchmark.
Since CLEF includes long documents beyond 512 tokens, we enable the maximum token limit for all models during evaluation on this dataset -- 512 tokens for E5, 8192 tokens for all other models.

\section{Replication of ``Language Gap'' with Fewer Pretraining Samples} \label{appendix:small-scale-ablation}

We perform a comparison of several variations of our training recipe to confirm our intuition that we are not observing any significant ``language gap'' in our training. In a shortened training procedure with only 13K contrastive pretraining steps, we vary the following factors:
\begin{itemize}[leftmargin=*]
    \item MLM base model variants: English-only (En-GTE) vs. multilingual (mGTE);
    \item Pretraining data: English-only (English portion of our pretraining data only) vs. multilingual (original data mix);
    \item Fine-tuning data: English-only (English portion of our fine-tuning data) vs. multilingual (English fine-tuning data plus non-English MIRACL training set).
\end{itemize}

\begin{table}
\centering
\begin{adjustbox}{width=\columnwidth}
\begin{tabular}{@{}c|c|c|ccccc@{}}
\toprule
\multirow{2}{*}{MLM}       & \multirow{2}{*}{PT} & \multirow{2}{*}{FT} & \multicolumn{4}{c}{Evaluation Sets} \\ \cmidrule(l){4-7} 
                             &                          &                         & \multicolumn{1}{c}{MTEB-R}  & \multicolumn{1}{c}{CLEF}  & \multicolumn{1}{c}{MIRACL} & \multicolumn{1}{c}{MIRACL-O} \\ \midrule
\multirow{2}{*}{En}          & \multirow{4}{*}{En}      & En                       & 0.526 & 0.327  & 0.114       & 0.271     \\
                             &                          & Mul                      & \textbf{0.532} & 0.340 & 0.268       & 0.361    \\ \cmidrule(l){1-1} \cmidrule(l){3-7} 
\multirow{4}{*}{Mul}         &                          & En                       & 0.524 & 0.439 & 0.486       & 0.478    \\
                             &                          & Mul                      & \textbf{0.532} & 0.442 & 0.588       & 0.517   \\ \cmidrule(l){2-2} \cmidrule(l){3-7} 
                             & \multirow{2}{*}{Mul}     & En                       & 0.525 & 0.451  & 0.530       & 0.521   \\
                             &                          & Mul                      & 0.529 & \textbf{0.452} & \textbf{0.594} & \textbf{0.538}   \\ \bottomrule
\end{tabular}
\end{adjustbox}
\caption{Impact of masked language modeling (MLM) base model, pretraining (PT), and finetuning (FT) data configurations with either English-only (En) or multilingual (Mul) content, showing their impact on downstream retrieval evaluations.}
\label{tab:en_or_mul_ablation}
\end{table}

As shown by the results tabulated in ~\Cref{tab:en_or_mul_ablation}, none of the multilingual treatments we attempted induce a sizable decrease in MTEB Retrieval score compared to the all-English baseline. In fact, we actually observe a slight positive effect from including the non-English MIRACL finetuning datasets even on the English MLM base model.

\end{document}